\documentclass[letterpaper, 10 pt, conference]{ieeeconf}  
\IEEEoverridecommandlockouts                              

\usepackage[utf8]{inputenc}
\usepackage{graphics} 
\usepackage{epsfig} %
\usepackage{mathptmx} 
\usepackage{times} 
\usepackage{amsmath} 
\usepackage{amssymb}  
\usepackage{graphicx}
\usepackage{caption}  
\usepackage{subcaption}
\usepackage{algpseudocode}
\usepackage{algorithm}
\usepackage{transparent}
\usepackage{dsfont}

\usepackage{hyperref}
\usepackage{graphicx}
\usepackage{enumerate}
\usepackage[usenames,dvipsnames]{color}
\usepackage{makecell}
\usepackage{bbm}

\usepackage{tikz}
\usetikzlibrary{shapes,arrows}
\usepackage{mathtools}

\newcommand{\norm}[1]{\left\lVert#1\right\rVert}
\DeclareMathOperator{\arctantwo}{arctan2}
\DeclareMathAlphabet{\mathcal}{OMS}{cmsy}{m}{n}

\input{mysymbol.sty}

\title{\LARGE \bf Inverse Optimal Planning for Air Traffic Control }
 \author{ Ekaterina Tolstaya$^1$, Alejandro Ribeiro$^1$, Vijay Kumar$^1$,  Ashish Kapoor$^2$
\thanks{This work is supported by grants NSF DGE-1321851 and ARL DCIST CRA W911NF-17-2-0181.}
\thanks{$^{1}$Department of Electrical and Systems Engineering, University of Pennsylvania, Philadelphia, PA 19104, USA {\tt\small [eig, aribeiro, kumar]@seas.upenn.edu}}
\thanks{$^{2}$Microsoft Corporation, Redmond, WA 98052, USA {\tt\small akapoor@microsoft.com}}
}

\begin{document}

\maketitle
\thispagestyle{empty}
\pagestyle{empty}

%
\begin{abstract}
We envision a system that concisely describes the rules of air traffic control, assists human operators and supports dense autonomous air traffic around commercial airports. We develop a method to learn the rules of air traffic control from real data as a cost function via maximum entropy inverse reinforcement learning. This cost function is used as a penalty for a search-based motion planning method that discretizes both the control and the state space. We illustrate the methodology by showing that our approach can learn to imitate the airport arrival routes and separation rules of dense commercial air traffic. The resulting trajectories are shown to be safe, feasible, and efficient.
\end{abstract}


\section{Introduction}
Air traffic controllers (ATC) must follow a complex set of regulations, including requirements on spacing between airplanes, weather restrictions, and airport-specific departure and arrival protocols. Additionally, experienced ATCs often formulate strategies that balance various demands that arise from the complex interplay between these factors. The demand for ATC services, which are already stretched thin, will further increase due to the rapid progress in the field of aerial robotics. 

We tackle the problem of building an Autonomous ATC that could significantly reduce the load on the human operators. In particular, we envision a system that concisely describes the rules of air traffic control and supports dense autonomous air traffic around commercial airports. There are several key challenges in building such an autonomous system. First, there are various deterministic and stochastic variables, such as traffic density, weather, regulatory requirements, and local geography. Often, there are multiple compliant ways, that can be qualitatively very different, to route air traffic. While a human operator can seamlessly optimize across these factors, it is not clear how an algorithmic system should weigh and jointly optimize across all these different criteria to mimic the choices of the human ATC. Secondly, most of the existing ATC services are developed for standard fixed-wing aircraft and helicopters. It is not clear how the current ATC system should be extended to novel aerial vehicles, such as micro UAVs and VTOL vehicles, which might have completely different dynamic behavior. Finally, from the algorithmic point of view, such a planning task requires optimization across multiple dynamic agents, which quickly becomes intractable as the number of vehicles increases.

Our novel approach combines search-based motion planning and inverse reinforcement leaning to address these challenges. The key technical insight is that, given a planner, we can learn a reward function that an ATC might be optimizing by leveraging aircraft traces available via the U.S. Federal Aviation Administration's Aircraft Situation Display to Industry feed. In particular, we learn the parameters of the reward function that correspond to how different factors are considered for trajectory selection. 

The resulting trajectories attempt to imitate real air-traffic and are shown to be safe and feasible. The learned cost functions are interpretable and can be compared to existing standard procedures. Additionally, the decoupling between the planner and the reward functions means that we can change the dynamics to those of a new aircraft or aerial robot and the system can adjust without retraining. Further, leveraging robotic path-planners also helps with computational challenges and eliminates the need for directly learning a control policy.


\subsection{Autonomous Air Traffic Control}

There are significant efforts to redesign the air traffic control system to enable autonomous decision making. One popular research direction is focused on conflict resolution to prevent vehicles from entering unsafe states. For example, \cite{prandini2011toward} aims to quantify the complexity of the current air traffic situation and provides a scheme for distributed control that ensures safety and eliminates the possibility of collisions. \cite{mahboubi2015autonomous} develops a Markov Decision Process-based system for resolving conflicts in flight plans. There has been a significant thrust in the development of air traffic simulators and interactive systems for experiments with control algorithms and human factors. \cite{tumer2007distributed} describes a simulator developed by NASA and an example of a reinforcement learning approach for an airplane's greedy optimization of arrival times. Rather than manually designing a reactive system that eliminates conflicts, we seek to learn how to generate feasible trajectories for open-loop control in dense air traffic. 

\subsection{Control using Artificial Potential Fields}

In mobile robot navigation, the penalty on unsafe states can be formalized as a spatial potential field. The artificial potential field is a popular method for efficient obstacle avoidance behaviors \cite{koren1991potential}. First, a potential field is generated based on known obstacles, and then the gradient of this potential field determines the direction of motion. One major drawback is the lack of guarantees for arrival at the origin or obstacle avoidance. Getting stuck in local minima of the potential functions is a risk \cite{koren1991potential}.
The evolutionary artificial potential function method has also been used for real-time robot path planning in the presence of dynamic obstacles \cite{870304}. 

Potential functions have been used extensively to model air traffic. \cite{ramamoorthy2004potential} proposes a potential-field based system that can encode the effects of factors such as weather and air traffic density to manage en-route traffic. \cite{xu2010air} focuses on the problem of re-routing of air traffic due to weather by generating potential functions and addresses the problems of local minima. 

These works use heuristics to avoid local minima in the potential field. They also neglect the problem of path planning in dense airspace around airports immediately prior to landing. We follow the approach of \cite{liu2018towards}, which  uses an artificial potential function as a penalty so that an optimal planner avoids potentially unsafe states. We extend this approach to the Dubins Airplane model and learn the potential function.

\subsection{Learning from Demonstrations}

Inverse reinforcement learning is the problem of using expert demonstrations to learn the reward function that an expert is maximizing. This reward function can then be used to determine a controller that imitates the trajectories of the expert. 
Approaches such as behavior cloning \cite{pomerleau1991efficient} seek to directly find a mapping from states to experts' actions, but this may generalize poorly to new situations. Other approaches include  maximum margin planning \cite{ratliff2006maximum} and feature expectation matching \cite{abbeel2004apprenticeship}, but these suffer from an ambiguity because one policy can be optimal for many different reward functions. We apply the maximum entropy inverse reinforcement learning algorithm described by \cite{ziebart2008maximum}. This approach has been extended to deep reward functions and policies \cite{wulfmeier2015maximum} and has been used in conjunction with planning for manipulation tasks \cite{finn2016guided}. 

\section{Planning using the Dubins Airplane Model}

We model the state of a single airplane using the Dubins airplane model, a four-dimensional system with a configuration space, $\mathcal{S} = \mathbb{R}^3 \times \lbrack - \pi, \pi \rbrack$, with $\bbs = (x,y,z,\phi)$, where $x$, $y$ and $z$ describe the coordinates of the airplane in three-dimensional Euclidean space, and $\phi \in [-\pi,\pi)$ is the bearing of the airplane relative to the $+x$ axis \cite{chitsaz2007time}. This airplane model assumes a fixed speed of $v$ in the $xy$-plane. 

The system is controlled through the first derivatives of altitude and bearing, $\dot{\phi}$ and $\dot{z}$, with control inputs denoted as $u_z$ and $u_\phi$, respectively. The system can be described as:
\begin{equation}
    \dot{\bbs} = \begin{bmatrix} \dot{x} \\ \dot{y} \\ \dot{z} \\ \dot{\phi} \end{bmatrix} =  \begin{bmatrix} v \cos \phi \\ v \sin \phi \\ u_z \\u_\phi \end{bmatrix}
\end{equation}
This model can be used to plan a trajectory from an initial state $\bbs_0 = (x_0, y_0, z_0, \phi_0)$ to a given a goal configuration $\bbs_g = (x_g, y_g, z_g, \phi_g)$. This information is given to each airplane by air traffic control when the airplane approaches an airport. For example,  $\bbs_g$ may be the location and orientation of an assigned runway. We assume that these assignments are provided prior to planning. The problem of planning a safe, minimum-length trajectory from the current state of the airplane to the goal can be formalized as follows:
\begin{align} \label{eq:hardtraj}
&\argmin_{\bbs(t)} \int_{t_0}^T \norm{ \dot{\bbs}(t) } dt \\
\text{s.t.  }  &\dot{\bbs}(t) = \big(\cos \phi, \sin\phi, u_z (t),u_\phi (t)\big) \nonumber \\
&  \bbs(t_0) = \bbs_0, \bbs(T) = \bbs_g \nonumber \\
& \bbs(t) \in \mathcal{S}^{safe}, \bbu(t) \in \mathcal{U}  \nonumber 
\end{align}
where $\mathcal{S}^{safe}$ is the set of safe states and $\mathcal{U}$ is the set of allowed control inputs. To enable learning the set of safe states $\mathcal{S}^{safe}$, we can reformulate \eqref{eq:hardtraj} to use a soft penalty on possibly unsafe states, following the approach of \cite{liu2018towards}, which uses hard constraints in addition to a potential function that guides the planned trajectory farther away from obstacles. The penalty term, $J\big(\bbs(t)\big)$, will be learned from data. For now, we express the new minimum path length planning problem as:
\begin{align} \label{eq:softtraj}
&\argmin_{\bbs(t)} \int_{t_0}^T  \Big(1 + J\big(\bbs(t)\big)\Big) \norm{\dot{\bbs}(t)} dt \\
\text{s.t.  }  &\dot{\bbs}(t) = \big(\cos \phi, \sin\phi, u_z (t),u_\phi (t)\big) \nonumber \\
&  \bbs(t_0) = \bbs_0, \bbs(T) = \bbs_g \nonumber \\
&  \bbu(t) \in \mathcal{U} \nonumber 
\end{align}
The motion cost of a trajectory, $\tau: \lbrack 0, \tau \rbrack \rightarrow \mathcal{S}$,  is the sum of the path length and the line integral of the penalty:
\begin{equation}
C(\tau) = \int_{t_0}^T \Big(1 + J\big(\bbs(t)\big)\Big)  \norm{\dot{\bbs}(t) } dt. 
\end{equation}

Solving the motion planning problem \eqref{eq:softtraj} requires searching the space of all feasible trajectories. The continuous state and time problem is intractable, so we follow the approach of \cite{liu2017search} by discretizing the system in time with an interval of $\Delta t = 30$ seconds. We also discretize the controls to obtain a set of fixed-time motion primitives. The bearing is chosen from the set $ u_\phi \in \lbrace -\Delta \phi, 0, \Delta \phi \rbrace$ and the altitude change is chosen from the set $ u_z \in \lbrace -\Delta z, 0, \Delta z \rbrace$. 
The motion primitives induce a discretization of states and enable the use of search-based motion methods for planning feasible and resolution-complete solutions \cite{liu2017search}.  While continuous states are denoted $\bbs \in \mathcal{S}$, we denote the discretized states as $\bar{\bbs} \in \mathcal{G}$, where $\mathcal{G}$ is a 4-dimensional grid, $\mathcal{G} \subset \mathcal{S}$. The induced discretization has a resolution of  $\bbrho = \lbrack \rho_x, \rho_y, \rho_z, \rho_\phi \rbrack$, so the conversion from $\bbs$ to $\bar{\bbs}$ can be expressed using the floor function $\lfloor \text{ } \rfloor$ :
\begin{equation} \label{eq:discretization}
\bar{\bbs} = 
\Bigg\lbrack
\bigg\lfloor \frac{\bbs_x}{\bbrho_x}\bigg\rfloor ,  
\bigg\lfloor \frac{\bbs_y}{\bbrho_y}\bigg\rfloor ,
\bigg\lfloor \frac{\bbs_z}{\bbrho_z}\bigg\rfloor ,
\bigg\lfloor \frac{\bbs_\phi}{\bbrho_\phi}\bigg\rfloor
\Bigg\rbrack
\end{equation}
To solve the discretized problem, we apply the Anytime Repairing A* (ARA*) algorithm for finite-time path planning  \cite{likhachev2004ara}. 
Please note the  $\epsilon$-suboptimality guarantee for the ARA* algorithm, which is expressed as:
\begin{equation}
C(\tau^*) \leq C(\tau)  \leq \epsilon C(\tau^*) 
\end{equation}
where $\epsilon \geq 1$ and $\tau^*$ is the optimal path and $C(\tau^*)$ is its cost.  
The ability to sample suboptimal trajectories allows us to visit and learn about a larger variety of states during inverse reinforcement learning. 
Given the continuous trajectory defined by a series control inputs, we then use trajectory refinement to produce smoother trajectories using spline interpolation  \cite{liu2017search}. 

To use the ARA* algorithm for motion planning for fixed-wing vehicles, we must use a heuristic that provides a lower bound on the distance to the goal. For the non-holonomic Dubins airplane model, there are existing methods for computing the optimal path length from a start state to a goal state, but they are expensive to compute and require the consideration of low-altitude and high-altitude cases \cite{chitsaz2007time}. To simplify this computation, we always assume the high altitude case, which allows adding a helical descent or ascent without worrying about collisions with the ground. Our heuristic may therefore inadmissible, but is much more computationally efficient. The computation of this heuristic is summarized in Algorithm \ref{algo_dubins}: First, we use the Dubins car model to compute the minimum path length from the start state to the goal state in the $xy$ plane. Following the approach of \cite{dubins1957curves}, to find the shortest length path, we check each of six types of Dubins Car paths, with each path consisting of three segments: Right (R), Straight (S) and Left (L)  \cite{dubins1957curves}.  Then, we compute the minimum time necessary for the ascent or descent. Then, if there is not enough time for the airplane to change altitude, we add helical sections to the trajectory to allow the airplane to descend in the $z$ dimension and return to the same position and bearing in the $xy$ plane. 

\begin{algorithm}[t]  %
     \textbf{Input:} Start $\bbs_0$, goal  $\bbs_g$, forward speed $v$, max rate of climb $\Delta z$, turning rate $\Delta \phi$ \\
    \textbf{Output:} Minimum path length from start to goal, $d_{min} $ 
\begin{algorithmic}[1]{


\State Compute Dubins car distance $d_{xy}$ using the LSL, RSR, LSR, RSL, RLR, LRL paths with curvature $\kappa = \Delta \phi / v$

\State Compute the minimum time from start to goal in the $xy$ plane
$$t_{min} = d_{xy} / v $$

\State Compute the minimum time for ascent/descent $$t_{z} = |z_g - z_0|/ \Delta z $$

\While {$t_{z} > t_{min}$}
\State Add a helical ascent/descent to the trajectory $$t_{min} = t_{min} + \frac{2 \pi}{\Delta \phi}$$
\EndWhile}

\State Compute the minimum path length from start to goal:
$$d_{min} = \sqrt{ (v \cdot t_{min})^2 + (z_g - z_0)^2  } $$
\end{algorithmic}
\caption{Dubins Airplane Heuristic}\label{algo_dubins}
\end{algorithm}


\section{Learning From Demonstrations}


We assume that the set of safe states $\mathcal{S}$ in \eqref{eq:hardtraj} is unknown. These conditions may be determined by the layout of the airspace at an airport, the motion of other airplanes, or weather. Our goal in this work is to learn about the unsafe conditions through inverse optimal control. To enable learning $J$ using gradient descent, we use the soft penalty formulation in \eqref{eq:softtraj}.

In order to learn the true penalty $J(\bbs)$, we require a data set  of expert demonstrations, $\mathcal{D}= \big\{ s_i^e(t) \big\}_{i=1,\ldots,M}$. Although we do not know the true penalty $J(\bbs)$, we have a current best estimate, $J_\theta(\bbs)$. This cost, $J_\theta$, is a function of non-linear features $\bbf$ of the state $\bbs$, defined in Section \ref{section_safety} along with particular examples of cost functions. We parametrize $J_\theta$ as linear in the features of the state $\bbf(\bbs)$:
\begin{equation}
    J_\theta (\bbs) = \theta^T \bbf(\bbs).
\end{equation}
The motion planner can use this estimated $J_\theta$ to plan trajectories satisfying \eqref{eq:softtraj}. Our next goal is to formulate the loss function that will allow us to update $J_\theta$ using trajectories from the expert and learner. 

We assume that the demonstrated expert trajectories follow the principal of maximum entropy  \cite{jaynes1957information}, which states that the probability of an expert's trajectory $\tau$ with a lower cost is exponentially more likely to be selected than a trajectory with a higher cost:
\begin{equation}
    \mathbb{P}(\tau) \propto e^{-C(\tau)}
\end{equation}
Using this assumption, we can apply the maximum entropy inverse reinforcement learning algorithm described by \cite{ziebart2008maximum}, and its deep learning counterpart \cite{wulfmeier2015maximum}. A detailed comparison of related works can be found in  \cite{finn2016guided}.  This approach seeks to find the cost function $J_\theta$ that \textit{maximizes} the log likelihood of expert trajectories $\mathcal{D}$:
\begin{equation}  \label{eq:objective}
   \mathcal{L}(\theta) = \log \mathbb{P}(\mathcal{D}, \theta | J_\theta) 
\end{equation}
If the cost function $J_\theta$ is a linear function the features, this problem is convex. \cite{ziebart2008maximum} shows that the gradient of this objective is the difference in feature counts along the trajectories of the expert and the learner. To compute the gradient of $\mathcal{L}$ with respect to $\theta$, we need to first introduce the state visitation counts of the expert computed using the data set $\mathcal{D}$:
\begin{equation}
    \bbf_\mathcal{D} = \sum_{ \tau \in \mathcal{D}} \sum_{\bbs \in \tau} \mathbb{P}(\tau)\bbf(\bbs)
\end{equation}
In \cite{ziebart2008maximum}, the expert's empirical feature counts are compared to the expectation of the learner's state visitation counts: 
\begin{equation}
    \mathbb{E}[\bbf_\theta] = \sum_{\bbs \in \mathcal{S} } \bbf(\bbs) \mathbb{P}(\bbs | J_\theta)
\end{equation}
Using these two quantities, we can express the gradient of $\mathcal{L}$ to be equal to the difference in feature counts along the trajectories of the expert and the learner \cite{ziebart2008maximum,wulfmeier2015maximum}:
\begin{equation}
     \nabla_\theta \mathcal{L} =  \mathbb{E}[\bbf_\theta] - \bbf_\mathcal{D}
\end{equation}
As in \cite{kalakrishnan2013learning}, we assume that the distribution of the learner's sampled trajectories is uniform. Therefore, the expectation $\mathbb{E}[\bbf_\theta]$ can be estimated using samples from learner with the current cost function $J_\theta$ and we can use the stochastic gradient computed using trajectories from the learner with the current cost $J_\theta$, $ \big\{ \bbs_i(t) \big\}_{i=1,\ldots,N}$, and trajectories from the expert, $\big\{ \bbs_i^e(t) \big\}_{i=1,\ldots,M}$ :
\begin{align} \label{eq:gradientstep}
 \hat{\nabla}_\theta \mathcal{L}=  \underbrace{ \frac{1}{N} \sum_{i=1}^{N} \;\; \sum_{t=t_0}^{T_i}  f\big(\bbs_i(t)\big)}_{\text{learner}} - \underbrace{\frac{1}{M} \sum_{i=1}^{M} \;\; \sum_{t=t_0}^{T_i} f\big(\bbs_i^e(t)\big)}_{\text{expert}}
\end{align} 
Gradient ascent on the objective $\mathcal{L}(J)$ will increase the cost of the states that the learner visits, but the expert does not, so that the learner will avoid those states in the future. In practice, we set $M=N=1$. 

We summarize our approach in Algorithm \ref{algo_ioc}. The learner obtains sets of time-synchronized expert trajectories for the landings of two or more airplanes. Then, for each trajectory in the expert's data set, the ARA* planner is used to plan a trajectory between the same start and end states.  Given the expert's and learner's trajectories, we can then compute the stochastic gradient and perform a gradient step on the $\theta$ parameter of the cost function. If the planner fails to produce a solution within the time limit, we use only the expert's trajectory for the gradient computation.

\begin{algorithm}[t]  %
\begin{algorithmic}[1]{
\For {$n=0,1,\ldots,M$}
\State Receive expert trajectories ordered by arrival time $$\mathcal{D} = \big\{ \bbs_i^e(t) \big\}_{i=1, \ldots, M}$$ 

\For {$i=0,1,\ldots,M$}
\State Obtain start state, $\bbs_0 $, and end state, $\bbs_g$, of $\bbs_i^e(t)$
\State Obtain other airplanes' trajectories, $\bbs_k^o(t)$ for $k<i$
\State Use ARA* planner to minimize $J_\theta$ over  $\bbs(t)$ 
\State Compute stochastic gradient:
$$ \hat{\nabla}_\theta \mathcal{L} =    \sum_{t=t_0}^{T}  f\big(\bbs(t)\big) -  \sum_{t=t_0}^{T_i} f\big(\bbs_i^e(t)\big)$$
\State Update cost parameters $\theta$  with step size $\alpha$:
$$\theta_{t+1} = \theta_t +  \alpha \hat{\nabla}_\theta \mathcal{L} $$
\EndFor
\EndFor}
\end{algorithmic}
\caption{Inverse Optimal Control for Air Traffic}\label{algo_ioc}
\end{algorithm}


\section{Safety in a Multi-agent System} \label{section_safety}

Our next goal is to use prior knowledge to add structure to the cost function $J$ to speed up learning. We assume that the centralized air traffic controller plans the landing trajectories for airplanes in the order of their arrival. When planning the trajectory for every new airplane, $\bbs(t)$, the trajectories of the $n$ previous airplanes are known, denoted $\big\lbrace \bbs_k^o(t) \big\rbrace_{k=1,\ldots,n}$. We also know the location of the destination airport, $\bbs_a$. 

Also, recall that the planning problem is computed in a discrete state space $\bar{\bbs} \in \mathcal{G}$, so we will only need to compute the cost of states on this discrete grid and can use the conversion from $\bbs$ to $\bar{\bbs}$ in \eqref{eq:discretization}. Therefore, rather than directly learning $J$, we only need to learn $\bar{J}: \mathcal{G}\rightarrow \mathbb{R}$. To relate this form to the previous notation, this is equivalent to choosing a feature extraction function $\bbf(s)$ that converts the continuous states $\bbs$ to discrete states $\bar{\bbs}$ and then applies additional non-linear operations. 

 We assume that $\bar{J}$ depends on two types of safety constraints: location relative to a specific airport, and the pairwise spacing between airplanes.  The first component of the cost function $\bar{J}$ controls the airspace around airport $a$ and can be expressed as $\bar{J}_a \big(\bar{\bbs}(t)\big)$. The second cost function controls the pairwise spacing of airplanes and can be written as $\bar{J}_o \big(\bar{\bbs}(t),\bar{\bbs}_k^o(t)\big)$, where $\bar{\bbs}_k^o(t)$ is the location of a nearby airplane at time $t$. If there are $n$ other airplanes in the area, we must sum this objective for all other airplanes: $\sum_{k=1}^{n} \bar{J}_o \big(\bar{\bbs}(t),\bar{\bbs}_k^o(t)\big)$.  Therefore, $\bar{J}\big(\bar{\bbs}(t)\big)$ can be written as a sum of the two types of soft penalties:
\begin{equation}
\bar{J}(\bbs(t)) = \bar{J}_a\big(\bar{\bbs}(t)\big) + \sum_{k=1}^{n} \bar{J}_o\big(\bar{\bbs}(t), \bar{\bbs}_k^o(t)\big)
\end{equation}
Next, we parametrize $\bar{J}_a$ and $\bar{J}_o$ to allow us to learn these functions using gradient descent. Motivated by planning in the presence of potential functions \cite{liu2018towards}, we choose to represent $\bar{J}_a$ as a spatial potential field, approximated using a fine discretization of the input space. This approximation represents the cost function as a value for each state $\bar{\bbs} \in \mathcal{G}$  on a discrete grid. A cost function used for motion planning must be non-negative. Also, the formulation must be piece-wise linear in features of state, $\bbf(\bbs)$, to satisfy the assumptions of \cite{ziebart2008maximum}. Therefore, we use a cost function of the following form:
\begin{equation} \label{eq:ja}
    \bar{J}_a \big(\bar{\bbs}(t)\big) = \sum_{\bar{\bbs}_i \in \mathcal{G}}  \max \big\{w_i  , 0 \big\} \mathds{1}_{\bar{\bbs}(t) = \bar{\bbs}_i}\big(\bar{\bbs}(t)\big),
\end{equation}
where $\mathcal{G} \subset \mathbb{R}^4$ is a finite set of all states within the controlled airspace and $\mathds{1}$ is the indicator function. By learning $w_i$ for each discrete state in the space, we construct a look-up table of costs to enable efficient motion planning. 

Next, we describe the penalty on pairwise distances between airplanes to control the airspace around each airplane. Each prior arrival is treated as a moving obstacle with a known trajectory, $\bar{\bbs}_k^o(t)$, and a cylindrical shape. We choose the penalty on getting too close to another airplane to be a linear drop-off potential function \cite{murphy2000introduction}:
\begin{align} \label{eq:jo}
\bar{J}_o \big(\bar{\bbs}(t), \bar{\bbs}_k^o (t)\big) = & u \max \big\{v_{z} - |\Delta_{z}(t)| ,0 \big\}   \nonumber \\ &\cdot  \max \big\{v_{xy} - \sqrt{ \Delta_x^2(t) + \Delta_y^2(t) } ,0 \big\}, 
\end{align}
where $\Delta(t) = \bar{\bbs}(t) - \bar{\bbs}_k^o(t)$ and $\Delta_x(t)$, $\Delta_y(t)$ and $\Delta_z(t)$ are the components of the difference in the discrete positions of the two airplanes at time $t$. 
We consider the relative positions in the horizontal and vertical dimensions separately since the clearance requirements in altitude may be different. The $v_{xy}$ and $v_z$ thresholds are learned through gradient descent, while the scaling factor $u$ is determined as a hyper parameter. The parameters for $\bar{J}_a$ and $\bar{J}_o$ can now be learned through stochastic gradient descent:
\begin{equation}
    \theta = \Big[\{w_i\}_{\bar{\bbs}_i \in \mathcal{G}}, v_{xy}, v_z \Big] 
\end{equation}
The number of  parameters $w_i$ is large and equal to $|\mathcal{G}|$. In the next section, we describe the practical considerations of learning this set of parameters. 

\section{Methods and System Architecture}

Now, we describe the implementation of the inverse optimal control system that learns from the air traffic data set.

\subsection{Dataset Processing}

\begin{figure}[t]
\begin{center}
{\transparent{0.7}\includegraphics[width=8cm]{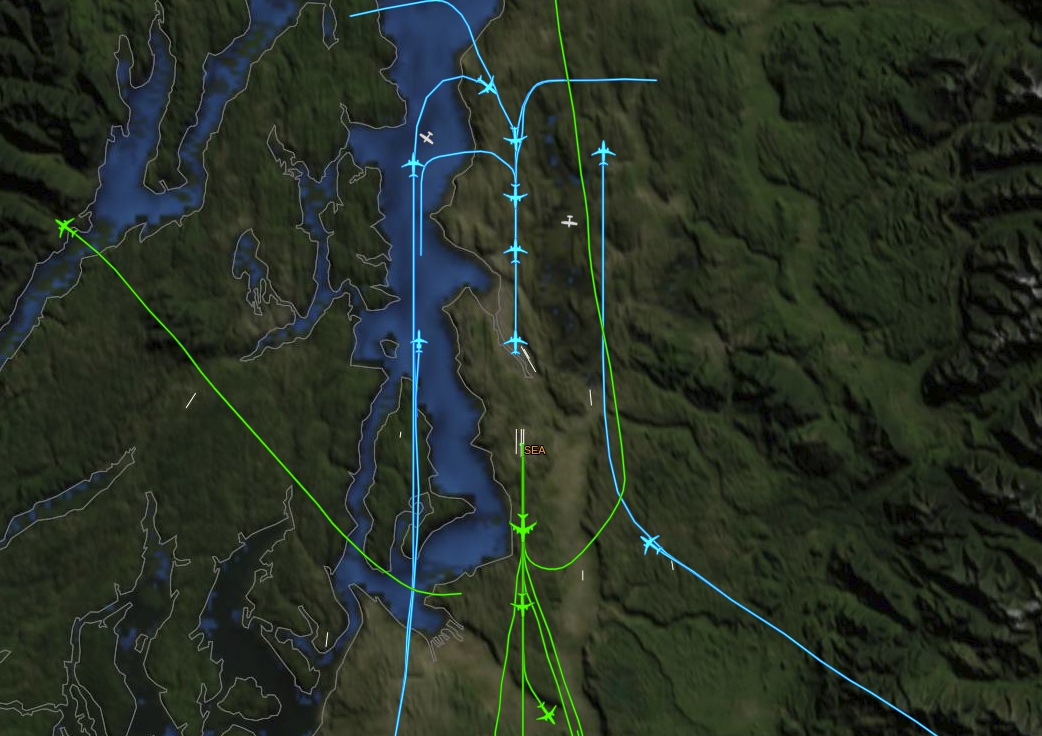}}
\end{center}
\caption{
A visualization of the Seattle-Tacoma airport data provided by FlightAware, with arrivals indicated in teal and departures in green \cite{flightaware}.}
\label{fig:seatac_flightaware}
\end{figure}

In this work, we narrow our focus to airplane landings at the Seattle-Tacoma airport (SEA), but this approach can be generalized to take-offs or inter-airport routing. Specifically, our goal is to mimic the Standard Terminal Arrival Routes at the SEA airport \cite{stars}.  We use recorded trajectories from the 11\textsuperscript{th} - 13\textsuperscript{th} of January, 2016 as provided by FlightAware \cite{flightaware}. A visualization of current data is displayed in Figure \ref{fig:seatac_flightaware}. The data set provides the location of airplanes asynchronously at a time interval of about 30 seconds. 

The airplane locations were provided as GPS measurements of latitude, longitude and altitude from onboard instrumentation. Bearing was estimated from subsequent observed locations. 
The provided WGS-84 measurement (GPS latitude, longtiude and altitude location) were converted to a local east-north-up (ENU) Cartesian system centered on the location of the Seattle-Tacoma airport \cite{farrell1999global, geodetictoenu}. All distances are provided in kilometers and bearing angles in radians from the $+x$ axis. The $+z$ axis indicates up and is perpendicular to the tangent plane.

By utilizing the Dubins airplane model, we make a constant velocity assumption. Real airplane trajectories accelerate and decelerate, especially during take-offs and landings. Extending our approach to acceleration or jerk-based motion primitives as in \cite{liu2018search} is left to future work.

\tikzstyle{block} = [rectangle, draw, fill=blue!20, 
    text width=6em, text centered, rounded corners, minimum height=4em]
\tikzstyle{blockred} = [rectangle, draw, fill=red!20, 
    text width=6em, text centered, rounded corners, minimum height=4em]
\tikzstyle{line} = [draw, -latex', <->]

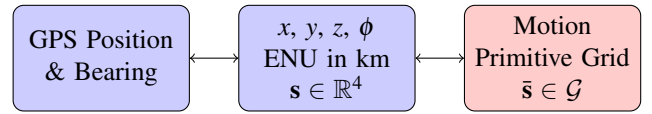
\begin{figure}[t]
\begin{center}
\begin{tikzpicture}[node distance = 3cm, auto]
    \node [block] (gps) {GPS Position \& Bearing};
    \node [block, right of=gps] (xyz) {$x$, $y$, $z$, $\phi$ ENU in km \\ $\bbs \in \mathbb{R}^4$};
    \node [blockred, right of=xyz] (prim) {Motion Primitive Grid \\ $\bar{\bbs} \in \mathcal{G}$};

    \path [line] (gps) -- (xyz);
    \path [line] (xyz) -- (prim);
\end{tikzpicture}
\end{center}
\caption{
Three parametrizations are used for expressing airplane states. The raw data provided by FlightAware is given in GPS latitude, longitude and altitude. We can convert from GPS to a local east-north-up (ENU) Cartesian frame. Then, we discretize the state with a resolution of $\rho$ to obtain the motion primitive grid coordinates via \eqref{eq:discretization}. Continuous state representations are denoted in blue and discrete - in red. 
}
\label{fig:seatac_satellite}
\end{figure}

\subsection{Interpolation}

The FlightAware data set provides waypoints at a time interval of about 30 seconds. We needed to perform interpolation, which is required for three reasons: 1) to generate time-synchronized trajectories for multiple airplanes, 2) to approximate the cost along a continuous trajectory, and 3) to generate dense trajectory data for SGD updates.  The interpolation is computed for the $x$, $y$ and $z$ components of the trajectory separately using the Univariate Spline module of the SciPy library \cite{spline}. The bearing of the aircraft is then computed using subsequent $x_t, y_t, z_t$ locations, $\phi_t = \arctantwo(y_{t+1} - y_t, x_{t+1} - x_t)$. 

\subsection{Planning }

The ARA* planner is given a fixed time limit of $30$ seconds. The planner does not always find a solution within the time limit, in which case we assume that the learner obtained a trajectory with zero cost and no states. This provides a lower bound on the learner's cost and assumes that the learner is able to instantly move directly from that start to the goal.  In this case, we can use only the expert's trajectory for the stochastic gradient to update the cost function.
The parameters of the motion primitives, such as the maximum turning rate and maximum rate of climb, were chosen based capabilities of commercial aircraft and tuned through a grid search procedure by comparing with actual trajectories. The parameters were chosen to be: forward velocity $v=100$ m/s, rate of climb $\Delta z = 6$ m/s, time discretization $\Delta t = 30$ s, angular velocity $\Delta \theta_1 = 0.025$ radians/s. An additional allowed turning rate of $\Delta \theta_2 = 0.0025$ radians/s mitigated some of the imprecision resulting from the coarse time discretization. The Dubins airplane heuristic still uses the largest turning speed to compute an approximate lower bound on the path length. The airplane states were discretized to a resolution of $\rho_x = \rho_y = 125$ m in the $x$ and $y$ dimensions and $\rho_z = 50$ m in the $z$ dimension. The bearing was discretized to $\rho_\phi = 0.05$ radians. 
Also, due to the discretization of the control space, it is extremely unlikely that the planned trajectories end up at exactly the goal state. Therefore, rather than having a single goal state, we use a goal region of  $\pm \lbrack 500, 500, 25 \rbrack$ meters and $\pm 0.125$ radians centered around the original goal state. 
\subsection{Learning }

To fit the cost function $\bar{J}_a$ as described in \eqref{eq:ja}, we need to store cost function values for each discrete state in the state space. We observe that the true structure of the data is sparse, so rather than storing a dense look-up table of cost values, we implement a sparse hash-map approach, which allows a constant-time evaluation of a state's cost. $\bar{J}_a$ is initialized to a cost of $w_i = 100$ for all states. When we observe a particular state and perform a gradient step, we store the new value in the hash table.  To further improve computational efficiency, the cost function was discretized on a slightly coarser grid that that of the motion primitive discretization, with resolution $0.25$ in the $x$ and $y$ dimensions and $0.125$ in the $z$ dimension. The bearing was discretized to $0.125$ radians. A step size of $\alpha = 10.0$ was used for learning.

The discretization of the cost function presents a challenge if the localization information is not exact. The provided dataset uses GPS locations, which are susceptible to drifts and sensor noise.
One possible solution is evaluating the cost function at planning time by computing the average of multiple nearby cells of the cost function grid. This is expensive due to the sparse data structure used to store the values of the cost function. Instead, we add Gaussian noise to the state $\bbs_t$ before the gradient update of the cost function \eqref{eq:gradientstep}. A small amount of white Gaussian noise is added to the input state $\bbs_t$ before discretization:  $\bbw_t \sim \mathcal{N}(0, \Sigma)$, where $\Sigma = \lbrack 0.25, 0.25, 0.125 \rbrack I$. The cost function $\bar{J}_o$ was initialized with overly conservative thresholds of $\lbrack v_{xy} = 60, v_{z} = 60\rbrack$ in discretized units, which is equivalent to $\lbrack 7.5, 3 \rbrack$ km. A step size of $\alpha=0.01$ was used, with gradients larger than $100.0$ clipped. The slope of the cost function is $u = 1.0$.

\begin{figure}[t]
\begin{center}
\includegraphics[width=7.5cm]{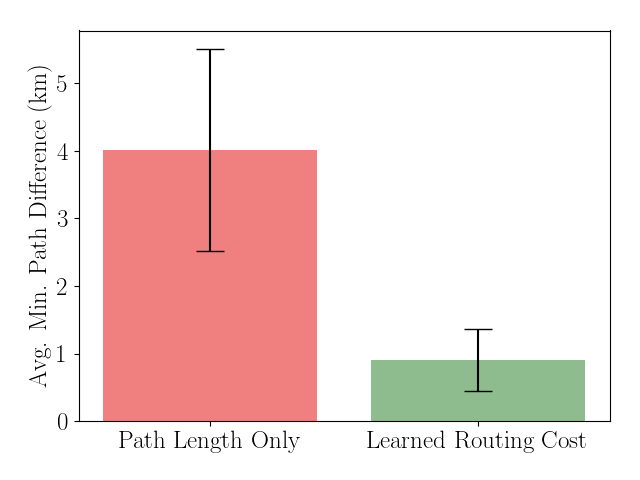}
\end{center}
\caption{
We quantify the planning performance of a planner that uses the learned $\bar{J}_a$ function and a planner that minimizes the path length criterion only. The average minimum path difference is $ \frac{1}{N} \sum_{i=1}^N \frac{1}{T} \sum_{t=1}^T \min_u ||\bar{\bbs}_i (t) - \bar{\bbs}_i^e(u)||_2 $, a measure of how far the learner's trajectory strays from the expert's demonstration on average. Error bars denote a one standard deviation bound and $N=300$. 
}
\label{fig:bad_good}
\end{figure}


\begin{figure*}[t]
\centering
    \begin{subfigure}{0.48\textwidth}
        \centering
        \includegraphics[width=\textwidth]{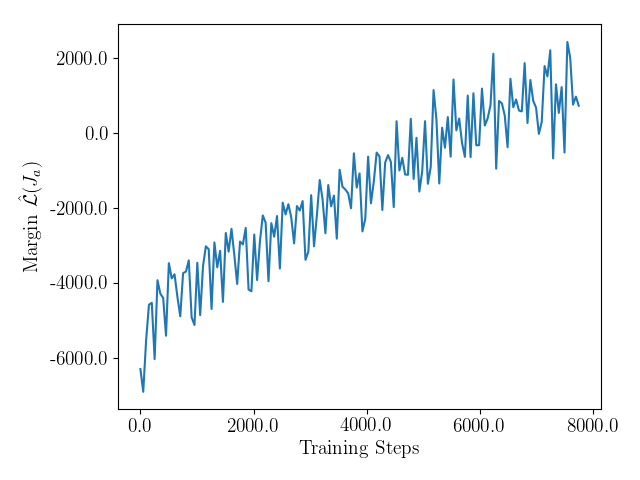}
        \caption{Cost Margin $\hat{\mathcal{L}}(\bar{J}_a)$}
        \label{margin}
    \end{subfigure}
    \begin{subfigure}{0.48\textwidth}
        \centering
        \includegraphics[width=\textwidth]{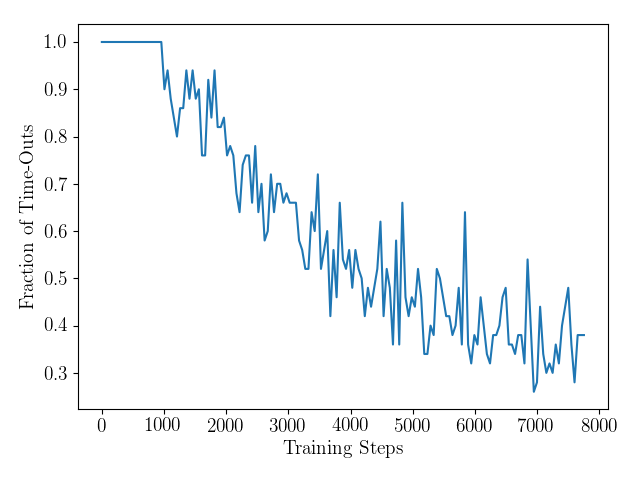}
        \caption{Fraction of Time-outs}
        \label{timeouts}
    \end{subfigure}
\caption{
First, we train the airport routing cost function $\bar{J}_a$. In Fig
\ref{margin}, we observe that, $\hat{\mathcal{L}}(\bar{J}_a)$, the margin between the expert's cost and the learner's cost increases during training, which we use as a proxy for the true optimization objective $\mathcal{L}(\bar{J}_a)$. 
The ARA* planner is given a fixed time budget and sometimes exceeds that time budget without finding a solution. Fig \ref{timeouts} shows the fraction of time-outs, which decreases during training. In both plots, each data point is the average of 50 consecutive training steps.
 }
\label{fig:margin_timeouts}
\end{figure*}

\begin{figure*}[t]
    \centering
    \begin{subfigure}{0.4\textwidth}
        \centering
        \includegraphics[width=\textwidth]{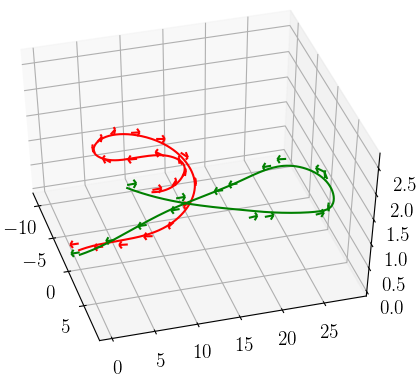}
        \caption{Minimum path length}
        \label{bad3d}
    \end{subfigure}
    \begin{subfigure}{0.4\textwidth}
        \centering
        \includegraphics[width=\textwidth]{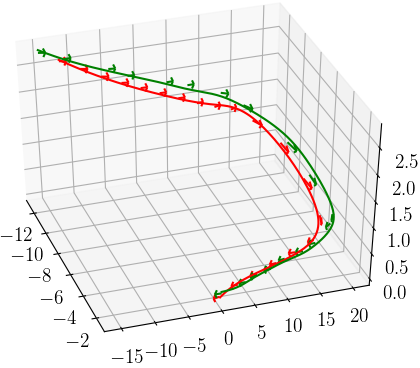}
        \caption{Using $\bar{J}_a$}
        \label{good3d}
    \end{subfigure}
\caption{
A comparison of the expert's and learner's 3D trajectories for two different objective functions. The actual ATC trajectory is denoted with green arrows and a green spline. The planner's path is marked with red arrows and interpolated by the red curve. The units are in kilometers in the ENU coordinate system. In \ref{bad3d}, the planner minimizes the path length alone, and in Figure \ref{good3d}, the planner uses the learned  airport routing cost, $\bar{J}_a$.   }
\label{fig:expert_and_learner_3d}
\end{figure*}


\begin{figure*}[t]
    \centering
    \begin{subfigure}{0.32\textwidth}
        \centering
        \includegraphics[width=\textwidth]{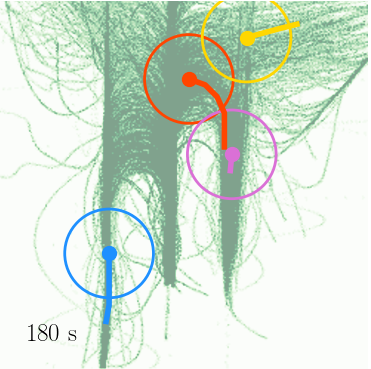}
        \caption{$t=180$ s}
        \label{time1}
    \end{subfigure}
    \begin{subfigure}{0.32\textwidth}
        \centering
        \includegraphics[width=\textwidth]{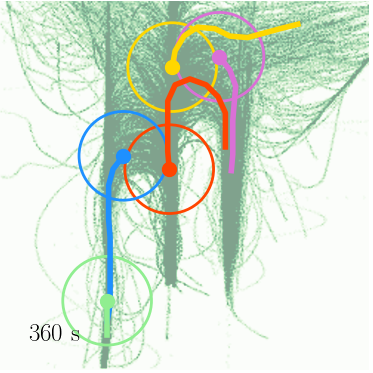}
        \caption{$t=360$ s}
        \label{time2}
    \end{subfigure}
        \begin{subfigure}{0.32\textwidth}
        \centering
        \includegraphics[width=\textwidth]{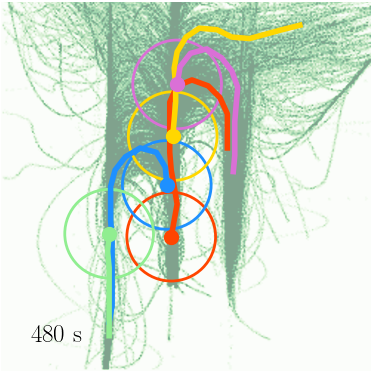}
        \caption{$t=480$ s}
        \label{time3}
    \end{subfigure}
    \caption{ A top down view of five trajectories produced by the learner using both $\bar{J}_a$ and $\bar{J}_o$, with the trajectories of the airplanes denoted in red, blue, yellow, purple and green, and their current locations denoted with a small filled circle. The large unfilled circles denote the threshold of the $\bar{J}_o$ cost, preventing airplanes from getting too close to each other. The learned cost function $\bar{J}_a$ is in the background indicated by green shading, with dark green indicating low cost states, and white indicating high cost states. We observe that the controller maximizes the efficiency of this landing sequence by pushing the trajectories of the airplanes as close together as allowed by the safety criterion.}
\label{fig:triple_landing}
\end{figure*}   


\section{Results}

The arrival route function $\bar{J}_a$ has an extremely large number of parameters and therefore requires more data to train. We trained this function first without considering distances between airplanes. Then, we fixed the routing cost function while learning the inter-airplane safety function $\bar{J}_o$.

\subsection{Learning the Airport Routing Cost}

First, we train the airport routing cost, $\bar{J}_a$, that defines the allowed paths for airplanes arriving to the Seattle-Tacoma airport. Rather than the expensive computation of $\mathcal{L}(J_\theta)$ \eqref{eq:objective}, we benchmark performance by the margin between the cost of the learner's trajectory and that of the expert, given an expert trajectory $\bar{\bbs}_i^e(t)$ and a corresponding planned trajectory $\bar{\bbs}(t)$: 
\begin{equation}
\hat{\mathcal{L}}(\bar{J}_a) =  \sum_{t=t_0}^{T_i}  \bar{J}_a\big(\bar{\bbs}(t)\big) -  \sum_{t=t_0}^{T_i} \bar{J}_a\big(\bar{\bbs}_i^e(t)\big)
\end{equation}
As the cost of the expert trajectory decreases relative to that of the expert, the objective increases and the probability of the expert's trajectory \eqref{eq:objective} increases. In Figure \ref{margin}, we observe that the benchmark $\hat{\mathcal{L}}(J_\theta)$ increases during training. 

It is important to note that the use of a fixed time limit for the ARA* planner often produces time-outs. In this case, the gradient step is computed using only the expert trajectory, treating the learner's trajectory as a cost of zero. Also, to speed up learning for the first 1000 iterations, we used only the expert's trajectory for the gradient step since the fraction of time-outs is large at the beginning of training.  It is interesting that the number of time-outs decreases during training, as we can see in Figure \ref{timeouts}. As the cost of the expert trajectories and the corresponding states decreases, the Dubins airplane heuristic becomes a tighter lower bound on the motion cost of the trajectories that move through those states. In Figure \ref{fig:expert_and_learner_3d}, we can see two examples of planned trajectories. We compare a trajectory planned using only the minimum path length objective to a trajectory planend using the learned $\bar{J}_a$ cost function. After training, the learner closely follows the expert's trajectory, with small oscillations in the z-axis. Figure \ref{fig:bad_good} quantifies this comparison  of the minimum path length planner and the routing cost planner over 1000 trajectories. 

\subsection{Learning the Inter-Airplane Safety Cost}

Now, we turn to learning the cost function that controls the spacing between pairs of airplanes, $\bar{J}_o$. We examine an example of five concurrent landing trajectories planned using the learned costs $\bar{J}_a$ and $\bar{J}_o$ in Figure \ref{fig:triple_landing}. The cost function $\bar{J}_a$ is visualized in green, with darker values indicating lower cost, and lighter values indicating higher cost. The five airplane trajectories are indicated with colored curves, with the current location of each airplane marked with a small filled circle. The thresholds of the $\bar{J}_a$ function are marked with large unfilled circles of corresponding colors. If a sixth airplane approached the airport, it would consider these unfilled circles to be states with very high cost and would avoid these regions, eliminating the posibility of collisions.

\section{Conclusion}

This work develops the method for search-based planning using cost functions learned through inverse optimal control. We demonstrate that the learned cost functions encode the implicit safety criteria of human ATC trajectories while maintaining high efficiency comparable to that of human operators. Our approach is well suited for planning trajectories over longer distances of tens of kilometers, but the Dubins airplane dynamics model has obvious limitations due to the coarse discretization of the control space,  which are especially apparent during the last stages of a landing.  Given a data set with more precise location measurements at a higher frequency, one possible solution could be to directly learn the motion primitives from data. This approach would still need to be verified in a high-fidelity air traffic simulation. 

We hypothesize that the learned cost functions could enable distributed control in dense airspaces, as an alternative to completely centralized trajectory generation and air traffic control. The airplane spacing function $\bar{J}_o$ learns the relative importance of other vehicles during trajectory planning, indicating which other vehicles can be ignored during planning. The airport routing function $\bar{J}_a$ may be found to encode which areas of the airspace are unused and can be utilized by other autonomous air traffic.

\bibliographystyle{IEEEtran}
\bibliography{IEEEfull,main}

\end{document}